\documentclass[lettersize,journal]{IEEEtran}
\usepackage{amsmath,amsfonts}
\usepackage{algorithmicx}
\usepackage{algorithm}
\usepackage{array}
\usepackage[caption=false,font=normalsize,labelfont=sf,textfont=sf]{subfig}
\usepackage{textcomp}
\usepackage{stfloats}
\usepackage{url}
\usepackage{verbatim}
\usepackage{graphicx}
\usepackage{cite}
\usepackage{stfloats}
\usepackage{amsthm,amsmath,amssymb}
\usepackage{mathrsfs}
\usepackage{algpseudocode}
\usepackage{amsfonts}
\usepackage{multirow}
\usepackage{float}
\usepackage{comment}
\usepackage{booktabs}

\begin{document}

\title{A Two-Stage Imaging Framework Combining CNN and Physics-Informed Neural Networks for Full-Inverse Tomography: A Case Study in Electrical Impedance Tomography (EIT)}

\author{Xuanxuan Yang$^{1,2}$ , Yangming Zhang$^{1}$, Haofeng Chen$^{1,2}$, Gang Ma$^{2}$, Xiaojie Wang$^{1}$

\begin{center}
    \vspace{10pt}
    {\footnotesize
    This is the author's accepted manuscript. The final published version will be available in IEEE Signal Processing Letters.
    
    \vspace{5pt}
    
    © 2025 IEEE. Personal use of this material is permitted. Permission from IEEE must be obtained for all other uses, including reprinting, republishing, or reuse of any copyrighted component of this work in other works.

    \vspace{5pt}
    
    The final published version is available at: [10.1109/LSP.2025.3545306]
    }
    
\end{center}
\thanks{XuanxuanYang, Haofeng Chen are with the Institute of Intelligent Machines, Hefei Institute of Physical Science and University of Science and Technology of China, Hefei. Gang Ma is with the University of Science and Technology of China, Hefei, Anhui 230026, China (e-mail: magang93@ustc.edu.cn).  Xiaojie Wang and Yangming Zhang are with the Institute of Intelligent Machines, Hefei Institute of Physical Science, Chinese Academy of Sciences, Hefei 230031, China (e-mail: xjwang@iamt.ac.cn; yz2898@columbia.edu).}
\thanks{This work was supported in part by the National Natural Science Foundation of China (Grant NO.62303436);Supported by Students ' Innovation and Entrepreneurship Foundation of USTC}
\thanks{Corresponding author: Gang Ma (e-mail: magang93@ustc.edu.cn)
Xiaojie Wang (e-mail: xjwang@iamt.ac.cn)}}

\markboth{© 2025 IEEE. Personal use of this material is permitted...
}%
{Shell \MakeLowercase{\textit{et al.}}: A Sample Article Using IEEEtran.cls for IEEE Journals}


\maketitle

\begin{abstract}
Electrical Impedance Tomography (EIT) is a highly ill-posed inverse problem, with the challenge of reconstructing internal conductivities using only boundary voltage measurements. Although Physics-Informed Neural Networks (PINNs) have shown potential in solving inverse problems, existing approaches are limited in their applicability to EIT, as they often rely on impractical prior knowledge and assumptions that cannot be satisfied in real-world scenarios. To address these limitations, we propose a two-stage hybrid learning framework that combines Convolutional Neural Networks (CNNs) and PINNs. This framework integrates data-driven and model-driven paradigms, blending supervised and unsupervised learning to reconstruct conductivity distributions while ensuring adherence to the underlying physical laws, thereby overcoming the constraints of existing methods. 

\end{abstract}


\section{Introduction}
\IEEEPARstart{E}{lectrical} Impedance Tomography (EIT) is an imaging technique used to reconstruct internal conductivity distributions by injecting currents through boundary electrodes and recording the resulting boundary voltage measurements \cite{somersalo1992existence}\cite{borcea2002electrical}\cite{hanke2003recent}. This process involves inferring internal structures from limited and often noisy external data, making it a highly complex and ill-posed inverse problem, where small measurement errors can lead to significant uncertainties in reconstruction \cite{kaipio2006statistical}.

Physics-Informed Neural Networks (PINNs) have recently emerged as a promising machine learning technique for solving partial differential equations (PDEs) by embedding physical laws directly into the loss function \cite{raissi2019physics,cuomo2022scientific}. This ability to enforce physical constraints during training allows PINNs to excel in approximating PDE solutions, demonstrating significant potential in applications such as computational fluid dynamics (CFD) and inverse problems \cite{cai2021physics}. Their capability to incorporate governing equations has further made them an appealing choice for tomographic imaging tasks, where maintaining physical consistency is crucial \cite{guo2023physics,ruan2024magnetic}. 

Bar et al. \cite{bar2021strong} were the first to systematically address the EIT problem using PINNs, demonstrating promising results. However, their work was limited to relatively simple reconstruction geometries and relied on restrictive assumptions that are impractical in real world scenarios. Building on this, Pokkunuru et al. \cite{pokkunuru2023improved} incorporated energy-based priors into PINNs to accelerate network convergence and enhance imaging accuracy. However, these methods face significant limitations in practical applications: 

\begin{enumerate}
\item Current PINN approaches rely on continuous internal potential data, which is unavailable in real applications that depend solely on boundary voltage measurements, limiting their practical effectiveness \cite{bar2021strong,pokkunuru2023improved}. 

\item  These frameworks require training K+1 networks for K excitation positions in EIT applications, drastically increasing computational cost and time\cite{bar2021strong,pokkunuru2023improved,guo2023physics}. 
\end{enumerate}

The primary motivation behind this work is to overcome the limitations of current PINNs in EIT and make them suitable for real-time sensor applications, where computational efficiency and fast decision-making are crucial. To overcome these limitations, we propose a two-stage hybrid learning framework combining the strengths of CNNs and PINNs. By integrating data-driven and model-driven methods, this approach ensures feasibility, efficiency, and physical consistency. 

The main contributions of this paper are as follows:

\begin{enumerate}
\item Hybrid Learning Framework: We propose a hybrid framework combining CNNs and PINNs for EIT, enhancing interpretability  compared to direct data-driven methods \cite{cuomo2022scientific}\cite{agnelli2020classification}\cite{fan2020solving}\cite{hamilton2018deep}\cite{hamilton2019beltrami}. The CNN extracts spatial features, while the PINN enforces physical laws and boundary conditions. 

\item Reduced computational complexity: By leveraging CNNs, our approach reduces the need for training K+1 networks to just two, significantly lowering computational costs while generating high-quality potential distributions for the EIT problem \cite{bar2021strong,cuomo2022scientific,pokkunuru2023improved}. 

\item Enhanced PINN Imaging Performance: Our method addresses the performance limitations of PINNs in practical EIT scenarios and demonstrates potential extensibility to other inverse imaging problems. 
\end{enumerate}

\section{Method}

We use a 16-electrode EIT system with a unit circle radius of 1, where the electrodes are evenly distributed along the boundary. The system operates with a complete electrode model (CEM) and follows a four-electrode measurement protocol. The resulting voltage data are then fed into our two-stage hybrid framework for image reconstruction, as illustrated in Figure 1. 

\begin{figure*}[hbtp]
\centering
\includegraphics[scale=1.1]{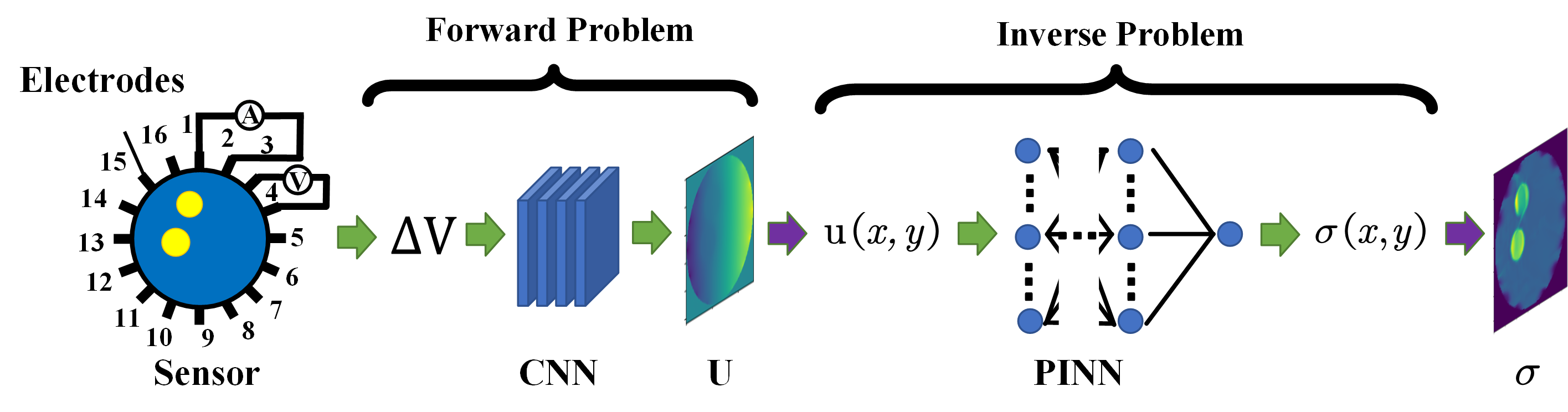}
\caption{The structural diagram of our proposed method.}
\label{fig_1}
\end{figure*}
\subsection{PDE in EIT}

The governing partial differential equation for EIT, derived from Maxwell's equations, is presented in Equation (1) \cite{cheney1999electrical}.The first term represents an elliptic equation, where $\sigma$ is the conductivity and $u$ is the electric potential. The second and third terms define the boundary conditions, with $n$ as the unit normal vector, $g$ as the current density, and $f$ as the measured potential. The fourth term represents the Neumann-to-Dirichlet (NtD) mapping. 

\begin{equation}
\label{deqn_ex1a}
\begin{cases}
	-\nabla \cdot \sigma \nabla u=0\quad \,\,                  \mathrm{in} \,\,\Omega\\
	\sigma \left( \frac{\partial u}{\partial n} \right) =g\,\, \,\,\,\,  \mathrm{on}\,\, \partial \Omega\,\,  \mathrm{Neumann}\,\, \mathrm{BC}\\
	u=f\,\,\,\,\,\,            \mathrm{on}\,\, \partial \Omega  \,\,\mathrm{Dirichlet}\,\, \mathrm{BC}\\
	\Lambda _{\sigma}: g\mapsto f\\
\end{cases}
\end{equation}

\subsection{Convolutional Neural Network}
As seen in Equation (1), solving the PDE requires knowledge of the electrical potential $u(x,y)$ and its derivatives. To address the challenges posed by data scarcity in EIT, our framework uses physics-driven synthetic training data generated through finite element simulations. This approach, commonly used in medical imaging, helps mitigate the issue of limited experimental data by augmenting the dataset with synthetic data\cite{shorten2019data}. In EIT, our synthetic data generation mimics the physical principles of current propagation and voltage responses, enabling the CNN to learn a generalized mapping from boundary measurements to internal potentials.

This hybrid method, combining physics-based simulations with data-driven learning, reduces reliance on experimental data while ensuring physical consistency. As a result, the model performs well even in data-scarce situations. While techniques such as using CNNs with CBAM attention mechanisms \cite{woo2018cbam}, residual networks, and auxiliary outputs of potential derivatives have shown reasonable success. Yet, when these outputs are fed into PINNs, training often fails due to the non-constant background, which lacks sufficient consistency for PINNs to function effectively. 

\begin{figure}[htbp]
\centering
\includegraphics[scale=1.2]{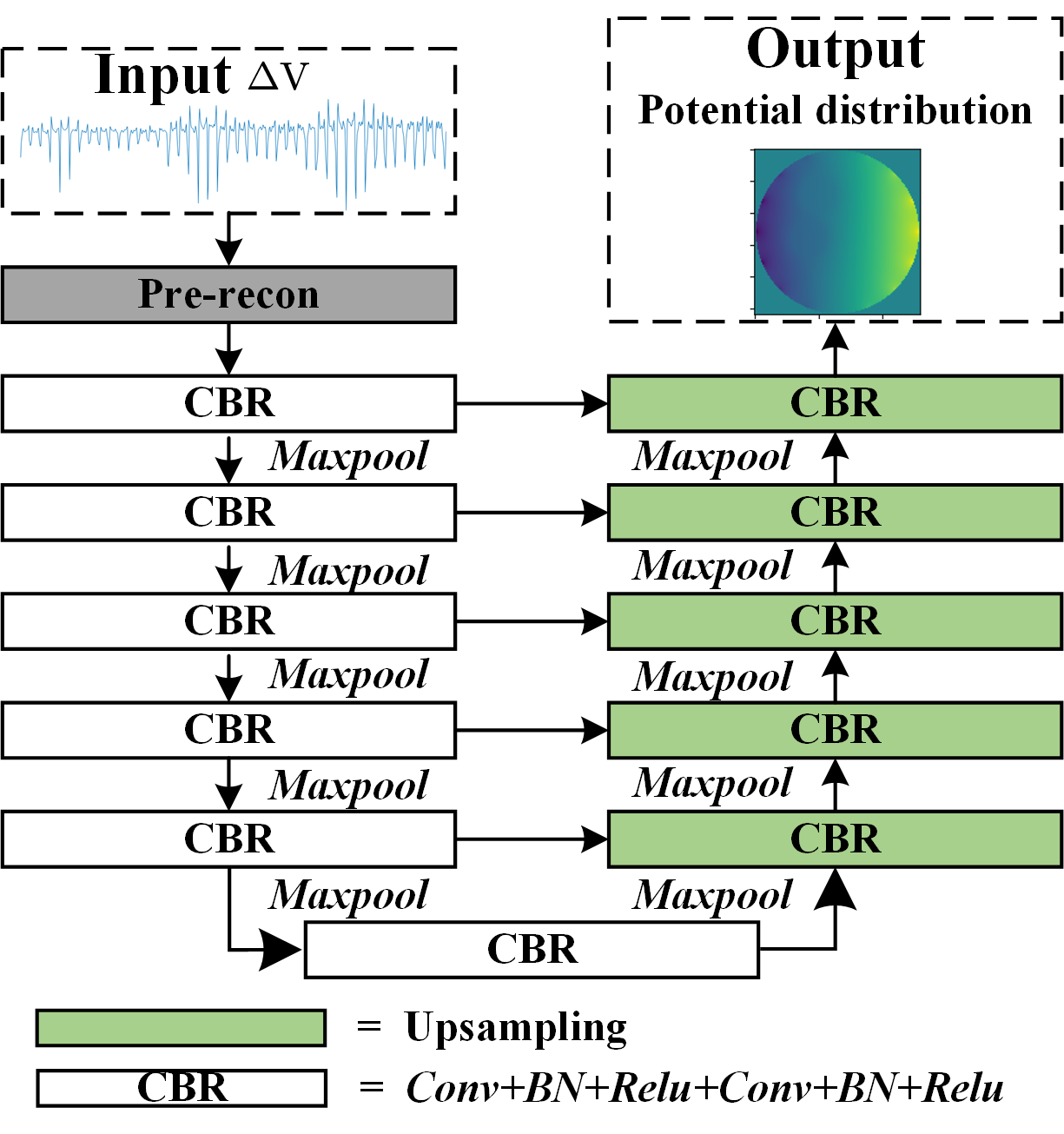}
\caption{The architecture of
U-Net.}
\label{U-Net}
\end{figure}

Inspired by the success of the U-Net architecture in image segmentation, particularly the design proposed by Qin et al. \cite{qin2020u2}, we employ a U-Net to establish an end-to-end mapping from the boundary voltage measurements $\Delta \mathrm{V}$ to the internal potential distribution $U$, as shown in Figure 2. Each CBR block within the U-Net consists of two convolutional layers, each followed by a batch normalization and a ReLU activation. The convolutional layers use 3×3 kernels, while max-pooling layers employ 2×2 kernels, ensuring consistent spatial scaling and efficient feature extraction. 

\begin{figure}[htbp]
\centering
\includegraphics[scale=0.8]{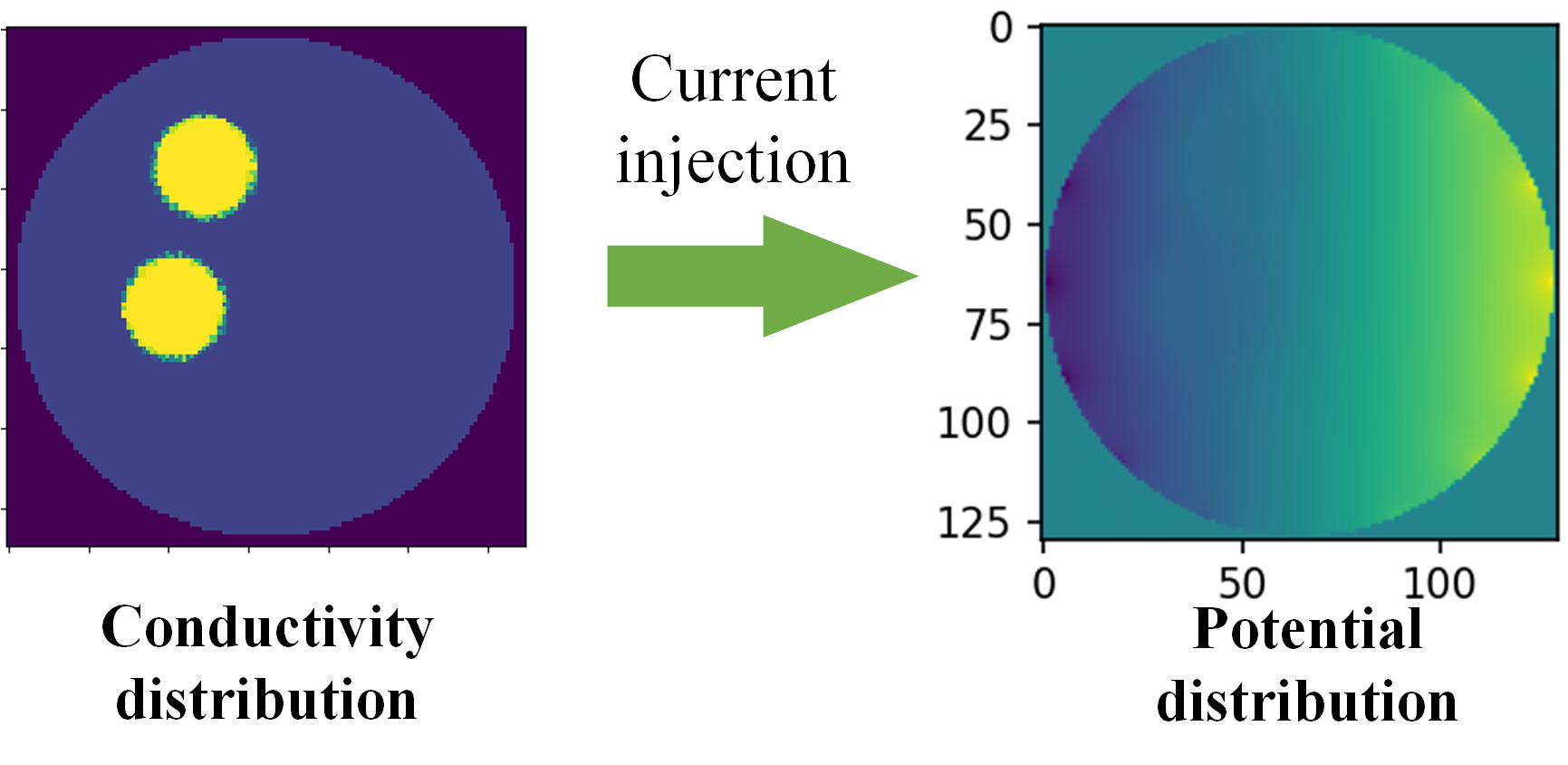}
\caption{Conductivity distribution inside the sensor and potential distribution formed after current injection.}
\label{fig_3}
\end{figure}

Building on related works \cite{bar2021strong,pokkunuru2023improved,siltanen2000implementation}, we define the current pattern as $g=\small{\frac{1}{\sqrt{2\pi}}}\sin \left( \small{\omega k+\varphi} \right) $, where $\omega =\small{\frac{2\pi}{16}}$ and $\varphi =0$ for the 16-electrode system. The resulting potential distribution is shown in Figure 3. We then train the network on a large dataset to predict the internal potential distribution based on boundary voltage measurements, using supervised learning to capture the spatial relationship between them.

Once the output $U$ is obtained from the CNN, the potential values $u(x,y)$ at any point within $U$ can be accessed, providing the necessary data to solve the PDE using PINN.

\subsection{Physics-Informed Neural Networks}

Inspired by Wang et al. \cite{wang2021understanding}, we use a PINN with a multilayer perceptron (MLP) consisting of four hidden layers, each with 64 neurons and tanh activation functions. To improve learning efficiency, we add residual connections at each layer. We tested architectures with 4, 6, and 8 hidden layers, finding that while more layers captured more complex relationships crucial for EIT, too many layers caused overfitting. Based on these experiments, we chose the 4-layer configuration. The input to the network comprises coordinates in the Cartesian domain $\Omega$, while the output represents the conductivity value $\sigma(x,y)$ at each spatial point.

\begin{equation}
\begin{aligned}
\mathcal{L} &=\frac{\alpha}{\Omega}\sum_{d\in \{\Omega \}}{\left( \nabla \cdot \left( \sigma _d\nabla u_d \right) \right)}^2\\
&+\frac{\beta}{M}\sum_{m\in top_M\mathcal{L} _{\mathrm{PDE}}}{\left| \nabla \cdot \left( \sigma _d\nabla u_d \right) \right|}\\
&+\frac{\gamma}{\left| \partial \Omega _b \right|}\sum_{b\in \partial \Omega _b}{\left| \sigma _b\frac{\partial u_b}{\partial n_b} \right|}+\frac{1}{|\partial \Omega |}\sum_{b\in \partial \Omega _b}{\left| \sigma _b-\sigma _{\partial \Omega _b}^{*} \right|}\\
&+\frac{\tau}{|\Omega |}\sum_{d\in \Omega}{\sqrt{\nabla _x\sigma _d+\nabla _y\sigma _d+\xi}}\\
&+\frac{v}{|\Omega \cup \partial \Omega |}\sum_{h\in \{\Omega \cup \partial \Omega \}}{\max \left( 0,1-\sigma _h \right)}+\zeta \left\| w_{\sigma} \right\| ^2
\end{aligned}
\end{equation}

To ensure physical consistency, the governing partial differential equations (PDEs) described in Equation (1) are incorporated into the MLP's loss function, resulting in the composite loss function presented in Equation (2). Given the ill-posed and highly nonlinear nature of inverse problems for elliptic PDEs, incorporating regularization methods is crucial for achieving a reasonable reconstruction \cite{jin2012analysis}\cite{jin2012reconstruction}. The loss function is composed of the following components: 

\begin{enumerate}
    \item \textbf{Elliptic equation loss}: The $L_2$ norm of the PDE residual, computed as the squared difference between the divergence of the conductivity-scaled gradient of the potential $\nabla \cdot \left( \sigma\nabla u \right)$ and its expected value, averaged over the entire domain. This term enforces the consistency of the potential $u(x, y)$ and conductivity $\sigma(x, y)$ with the elliptic equation. 
    \item \textbf{Top-$M$ norm loss}: An $L_\infty$ norm, which emphasizes the largest $M$ terms in the $L_2$ loss. By focusing on regions with the most significant errors, this term improves model robustness in critical areas. 
    \item \textbf{Boundary condition losses}: The next two losses enforce adherence to the Neumann and Dirichlet boundary conditions. Specifically, the Neumann loss accounts for the flux at the boundaries, while the Dirichlet loss constrains the potential to measured values at specific boundary points. Here, $\sigma _{\partial \Omega _b}^{*}$ represents the known conductivity values at the boundary. 
    \item \textbf{Total variation regularization loss}: Inspired by the work of Gonz{\'a}lez et al.\cite{gonzalez2017isotropic}, this isotropic total variation term acts as a regularizer, leveraging the prior knowledge that conductivity distributions are often sparse in their gradient domain. This ensures smoothness in the predicted conductivity map. 
    \item \textbf{Hinge loss}: To maintain physical validity, this term penalizes any predictions of conductivity values less than or equal to zero, as such values are non-physical in practical scenarios. 
    \item \textbf{Parameter loss}: This term mitigates overfitting, reduces the model’s sensitivity to noise or outliers, and enhances its ability to generalize to unseen data. 
\end{enumerate}

While Bar et al. \cite{bar2021strong} and Pokkunuru et al. \cite{pokkunuru2023improved} also employ PINNs in EIT, their methodologies rely on TensorFlow’s $\texttt{tf.gradients}$ function for automatic differentiation of the potential $u(x, y)$. Specifically, their approach involves training K networks to predict the potential $u(x, y)$ alongside a single network for the conductivity $\sigma(x, y)$. During training, $u(x,y)$ is updated based on $\sigma(x, y)$, and vice versa. However, this simultaneous training strategy has significant limitations. Without sufficient prior information, the network often fails to converge to meaningful solutions, resulting in poor reconstruction performance. 

In contrast, our approach employs a matrix-based method to compute the partial derivatives of the potential $U$ with respect to $x$ and $y$. By numerically differentiating the potential matrix $U$ via finite differences, we avoid the instability issues associated with simultaneous network training. This matrix-based method has demonstrated accuracy and stability, allowing for more reliable predictions of the conductivity map. However, it is worth noting that this method comes with a trade-off: the finite difference computation can slow down the overall convergence speed during training. 

\section{Experiments and results}

\subsection{Sensitivity Analysis}

The EIT problem's highly nonlinear and ill-posed nature makes it sensitive to parameter variations, where minor adjustments can significantly impact the reconstruction quality. To optimize performance, we systematically tuned the loss penalties and learning rate, which are key factors influencing the solution quality.

The hyperparameters in Equation (2) were selected from predefined ranges: $\alpha$$\in$$\{0.01, 0.05, 0.1, 0.5, 1\}$, 
$\beta$$\in$$\{0.01, 0.05,$\\$ 0.1, 0.5, 1\}$, $\gamma$$\in$$\{0.1, 0.5, 1, 1.5, 2\}$, $v$$\in$$\{5, 8, 10, 50, 100\}$, and learning rate $\in$$\{0.0001, 0.001, 0.01, 0.1\}$. The learning rate was adjusted dynamically using an exponential decay schedule with a decay rate of $0.9$ applied every $200$ epochs.

For each training run, we randomly sampled a single value from each hyper-parameter's search space to evaluate its effect on reconstruction quality. The final optimal hyper-parameters were determined and are summarized in \textbf{Table I}.

\subsection{Dataset}

\begin{figure}[hbp]
\includegraphics[scale=1.1]{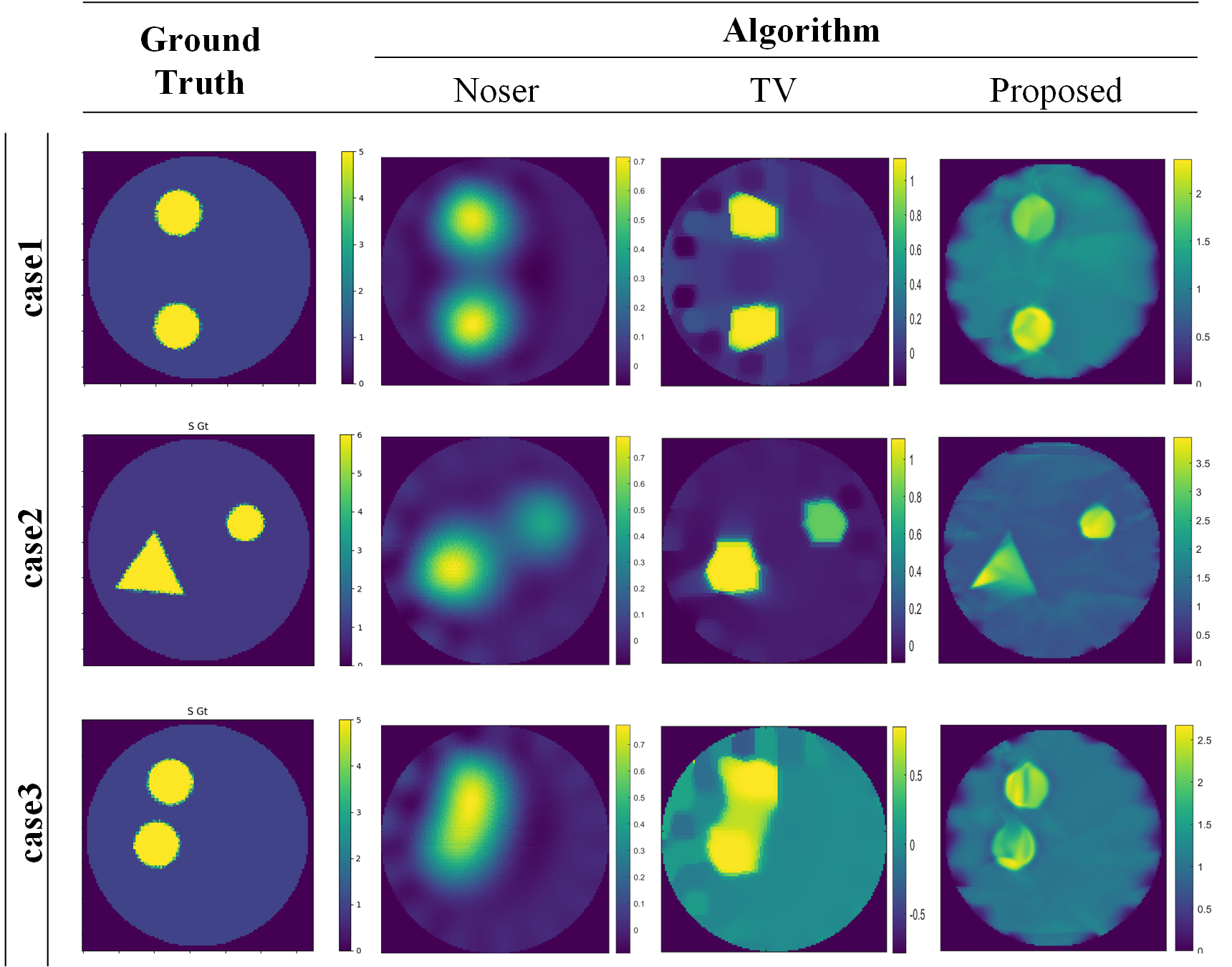}
\caption{Comparison of reconstructions.}
\label{fig_5}
\end{figure}

The dataset used to train the convolutional neural network consisted of 10,000 samples, with 10\% allocated for validation. To enhance robustness and further reduce the risk of inverse crime, an independently generated testing dataset was used, completely separate from both the training and validation sets. Additionally, this testing dataset was augmented with Gaussian noise at a signal-to-noise ratio of 20 dB to better simulate real-world measurement conditions. 

The training data was generated using MATLAB. We predefined configurations featuring a single circle, two circles, and a combination of a circle and a triangle. The background conductivity was fixed at 1, while the conductivity of the objects varied randomly between 2 and 6. To ensure diversity and robustness in model training and evaluation, the positions of the objects were randomized. Using these configurations, boundary voltage measurements under the current pattern $g$ were computed using EIDORS. 

For training, the input data comprised measured voltage differences, $\Delta \mathrm{V}$, while the output corresponded to the internal potential distributions calculated under the applied current pattern $g$. The training process was conducted on two NVIDIA GTX 3090 GPUs, taking approximately 5 hours to complete around 800 epochs. 

\subsection{Simulation experiments}
To evaluate the imaging performance, we present three distinct test results, each corresponding to a unique configuration with varying object shapes and conductivity patterns, as shown in Figure 4. These test cases were processed using our CNN-PINN networks, which allowed us to assess the network's capability to predict internal potential distributions and reconstruct conductivity from boundary voltage measurements.

Due to the lack of open-source access to Bar et al.'s work, we were unable to replicate their results. Instead, we compared our method with alternative approaches, as illustrated in Figure 4. The measurement metrics are summarized and further analyzed in Table II, providing additional insights into the performance of our approach. To quantitatively evaluate the performance of each method, we used three metrics: the Structural Similarity Index (SSIM), the Correlation Coefficient (CC), and the Relative Error Index (RIE). A higher SSIM value indicates greater similarity between the reconstructed and true conductivity distributions. The CC measures the linear relationship between the reconstructed and actual conductivity distributions, with a higher CC indicating a closer match. The RIE quantifies the relative error in the reconstruction, where a higher RIE value reflects a larger discrepancy between the reconstructed and true distributions.

Each training session, which involved processing multiple test cases, took approximately one hour on two GTX 3090 GPUs. The results demonstrate that our proposed method performs robustly across a variety of complex shapes and configurations. Even in the presence of challenging geometries, such as closely situated shapes or irregular conductivity distributions, our method successfully reconstructed the internal conductivity distribution, effectively addressing the EIT problem.

\begin{table}[h!]
    \caption{Assigned values of Hyper-parameters.}
    \centering
    \setlength{\tabcolsep}{4pt}
    \begin{tabular}{@{}c*{10}{c}@{}}
        \toprule
        \multicolumn{8}{c}{\textbf{Hyper parameters}} \\ 
        \midrule
                                   & $\alpha$ & $\beta$ & $M$ & $\gamma$  & $\tau$ & $\xi$ & $v$  & $\zeta$\\ 
                                   & 0.01 & 0.01 & 40   & 1.5  & 1    &  1e-4 & 8 & 1e-6\\
        \bottomrule    
    \end{tabular}
    \label{hyperparameters}
\end{table}

\begin{table}[h!]
    \caption{Evaluation metrics for Cases 1 to 3.}
    \centering
    \begin{tabular}{ccccccccc}
    \toprule
    \textbf{Case} & \textbf{Algorithm} & \textbf{SSIM $\uparrow$} & \textbf{CC $\uparrow$} & \textbf{RIE $\downarrow$} \\
    \midrule
    \multirow{2}{*}{Case 1} & NOSER    & 0.3645 & 0.6534 & 0.9465 \\
                       & TV       & 0.3829 & 0.6347 & 1.1034 \\
                       & Proposed & \textbf{0.4066} & \textbf{0.7043} & \textbf{0.9365} \\
    \midrule
    \multirow{2}{*}{Case 2} & NOSER    & 0.1943 & 0.3347 & 3.0641 \\
                       & TV       & 0.2161 & 0.2803 & 3.4514 \\
                       & Proposed & \textbf{0.3671} & \textbf{0.403} & \textbf{2.4903} \\
    \midrule
    \multirow{2}{*}{Case 3} & NOSER    & 0.3841 & 0.7071 & 0.6107 \\
                       & TV       & 0.2873 & 0.6148 & 0.9701 \\
                       & Proposed & \textbf{0.4036} & \textbf{0.7146} & \textbf{0.4954} \\
    \bottomrule
    \end{tabular}
\end{table}

\section{Conclusion and future work}

In this study, we propose a two-stage method that integrates Convolutional Neural Networks (CNNs) and Physics-Informed Neural Networks (PINNs) to improve the performance of PINNs in inverse tomography, using Electrical Impedance Tomography (EIT) as a case study. Our approach enhances the applicability of PINNs to inverse tomography and holds potential for extension to other inverse imaging problems. While this work addresses the limitations of PINNs by alleviating the need for unrealistic information, future efforts will focus on integrating the CNN and PINN into a unified model, ultimately enabling true real-time sensor applications for dynamic environments.

\clearpage
\bibliographystyle{IEEEtran}

\bibliography{bare_jrnl_new_sample4.bbl}

\vfill

\end{document}